\documentclass[letterpaper, 10 pt, journal, twoside]{IEEEtran}

\usepackage[T1]{fontenc}
\usepackage{times}
\usepackage{amsmath}
\usepackage{amssymb}
\usepackage{graphicx}
\usepackage[ruled,linesnumbered]{algorithm2e}
\usepackage{booktabs}
\usepackage{float}
\usepackage{bigstrut}
\usepackage{multirow}
\usepackage{balance}
\usepackage[hidelinks=true,bookmarks=FWLse]{hyperref}
\usepackage{caption}
\usepackage{subcaption}
\usepackage{lineno}
\usepackage{amsopn}
\usepackage{comment}
\usepackage{color}
\usepackage{cite}
\usepackage{algorithmic}
\usepackage{tabularx}
\usepackage{xcolor}

\makeatletter

\newcommand{\Rmnum}[1]{\expandafter\@slowromancap\romannumeral #1@}
\makeatother

\newcolumntype{L}[1]{>{\raggedright\arraybackslash}p{#1}}
\newcolumntype{C}[1]{>{\centering\arraybackslash}p{#1}}
\newcolumntype{R}[1]{>{\raggedleft\arraybackslash}p{#1}}

\title{Self-Supervised Drivable Area and Road\\ Anomaly Segmentation using RGB-D Data\\ for Robotic Wheelchairs}

\IEEEaftertitletext{\vspace{-1.5\baselineskip}}

\author{Hengli Wang, Yuxiang Sun and Ming Liu, \IEEEmembership{Senior Member, IEEE}
\thanks{Manuscript received: Feb, 24, 2019; Revised May, 23, 2019; Accepted July, 26, 2019.}
\thanks{This paper was recommended for publication by Editor Dan Popa upon evaluation of the Associate Editor and Reviewers' comments. This work was supported by the National Natural Science Foundation of China (Grant No. U1713211), the Research Grant Council of Hong Kong SAR Government, China, under Project No. 11210017, and No. 21202816, awarded to Prof. Ming Liu. \textit{(Corresponding author: Ming Liu.)}}
\thanks{The authors are with the Department of Electronic and Computer Engineering, The Hong Kong University of Science and Technology, Clear Water Bay, Kowloon, Hong Kong SAR, China (email: {\tt\footnotesize{hwangdf@ust.hk}}; {\tt\footnotesize{sun.yuxiang@outlook.com}}, {\tt\footnotesize{eeyxsun@ust.hk}}; {\tt\footnotesize{eelium@ust.hk}}).}%
\thanks{Digital Object Identifier (DOI): see top of this page.}
}

\begin{document}

\maketitle

\begin{abstract}
The segmentation of drivable areas and road anomalies are critical capabilities to achieve autonomous navigation for robotic wheelchairs. The recent progress of semantic segmentation using deep learning techniques has presented effective results. However, the acquisition of large-scale datasets with hand-labeled ground truth is time-consuming and labor-intensive, making the deep learning-based methods often hard to implement in practice. We contribute to the solution of this problem for the task of drivable area and road anomaly segmentation by proposing a self-supervised learning approach. We develop a pipeline that can automatically generate segmentation labels for drivable areas and road anomalies. Then, we train RGB-D data-based semantic segmentation neural networks and get predicted labels. Experimental results show that our proposed automatic labeling pipeline achieves an impressive speed-up compared to manual labeling. In addition, our proposed self-supervised approach exhibits more robust and accurate results than the state-of-the-art traditional algorithms as well as the state-of-the-art self-supervised algorithms.
\end{abstract}


\section{Introduction}

\IEEEPARstart{R}{obotic} wheelchairs are designed to improve the life quality of the disabled or elderly people by increasing their mobility. To this end, autonomous navigation has been intensively studied and become an essential capability for robotic wheelchairs. The segmentation of drivable areas and road anomalies refers to pixel-wisely identifying the areas and anomalies in images. It is a crucial component for autonomous navigation. Without correctly segmenting drivable areas and road anomalies, robotic wheelchairs could bump or even roll over when passing through road anomalies, which may cause injuries to human riders. In this paper, we define the drivable area as the area where robotic wheelchairs can pass through regardless of their sizes, while the road anomaly is defined as the area with the height larger than $5$cm from the surface of the drivable area. The segmentation of drivable areas and road anomalies could be addressed using semantic segmentation techniques.

RGB-D cameras, such as Kinect \cite{kinect}, are visual sensors that can stream RGB and depth images at the same time \cite{sun2019active, sun2018motion, sun2017improving}. We use an RGB-D camera for the segmentation of drivable areas and road anomalies in this paper. The reason why we use RGB-D camera is that the depth difference between road anomalies and drivable areas could be useful to distinguish them. Recent development of deep learning techniques has brought significant improvements on the semantic segmentation using RGB-D cameras \cite{fusenet, depthaware, lateef2019survey}. In order to train a deep neural network, we usually need a large-scale dataset with hand-labeled ground truth. However, generating such a dataset is time-consuming and labor-intensive. In order to provide a solution for the excessive consumption of time and labor for manual labeling, we present a self-supervised approach to segment drivable areas and road anomalies for robotic wheelchairs with an Intel Realsense D415 RGB-D camera. Fig. 1 shows our robotic wheelchair, RGB-D camera and mini PC. The Intel Realsense D415 is an active stereo camera that estimates distances with emitted infrared lights.

\begin{figure}[t]
   \centering
   \includegraphics[width=\linewidth]{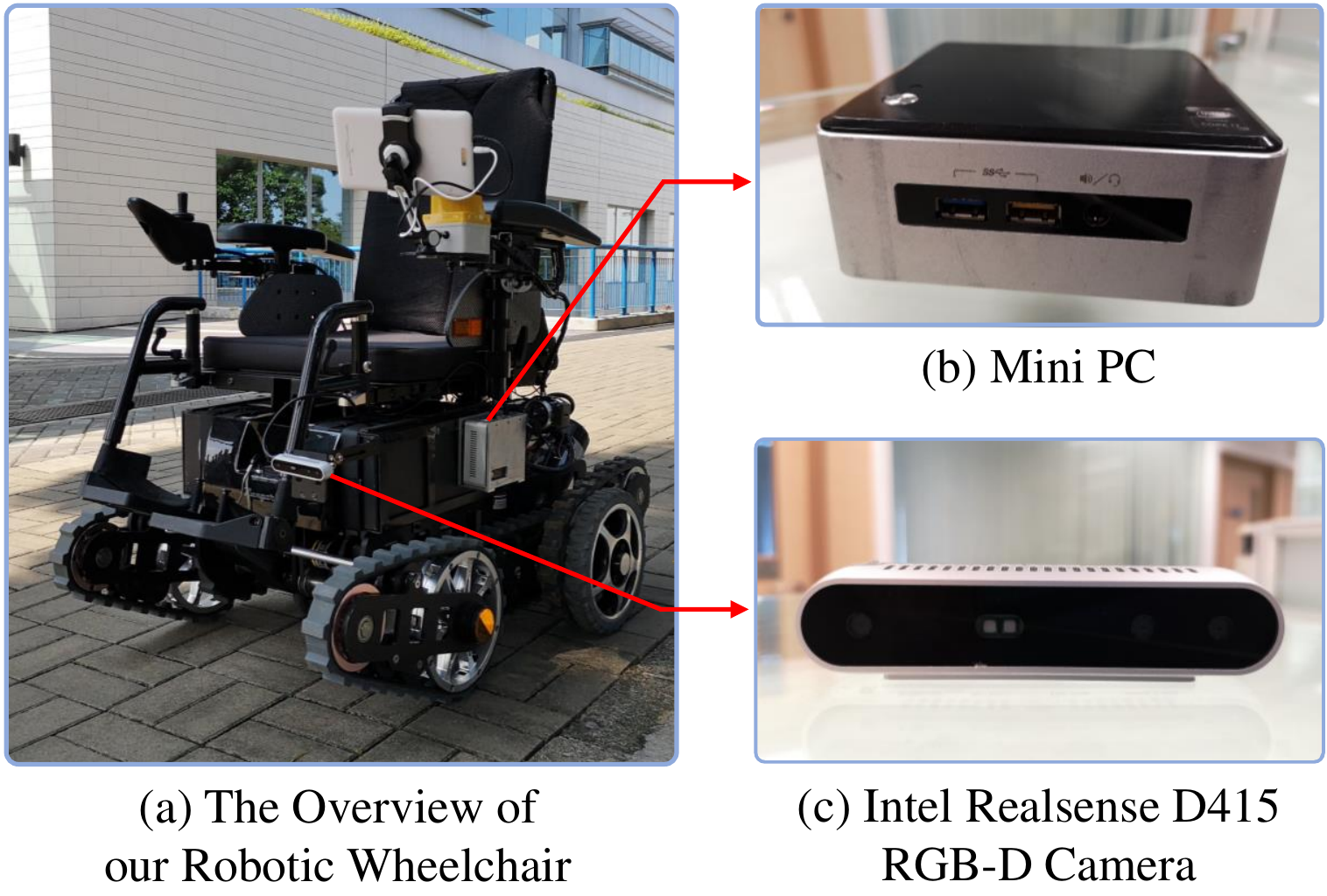}
   \caption{The robotic wheelchair used in this work. It is equipped with an Intel Realsense D415 RGB-D camera to collect data and a Mini PC to process data.}
   \label{wheelchair}
\end{figure}

\begin{figure}[tbp]
   \centering
   \includegraphics[width=\linewidth]{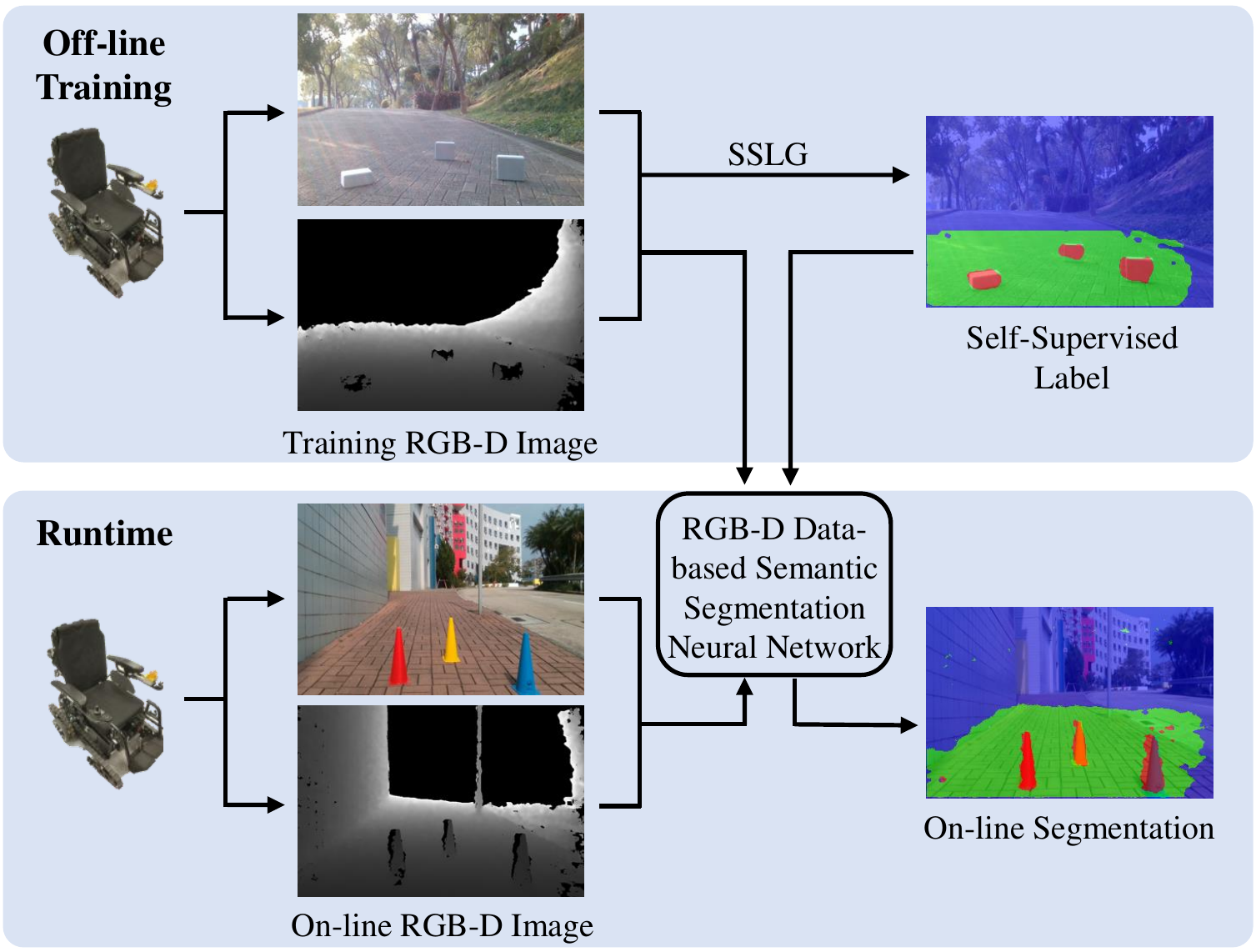}
   \caption{The overview of our proposed self-supervised approach for segmenting drivable areas and road anomalies. We firstly use our proposed SSLG to generate self-supervised labels (top), which are then used to train the RGB-D data-based semantic segmentation neural network. At runtime, a robotic wheelchair equipped with an RGB-D camera can perform on-line segmentation of drivable areas and road anomalies (bottom). The figure is best viewed in color.}
   \label{framework}
\end{figure}

Fig. 2 illustrates the overview of our proposed self-supervised approach. We firstly develop a pipeline named Self-Supervised Label Generator (SSLG) to automatically label drivable areas and road anomalies. Then, we use the segmentation labels generated by the SSLG to train several RGB-D data-based semantic segmentation neural networks. Experimental results show that although the segmentation labels generated by the SSLG present mis-labelings, the results of our proposed self-supervised approach still outperforms the state-of-the-art traditional algorithms and the state-of-the-art self-supervised algorithms. The contributions of this paper are summarized as follows:

\begin{enumerate}
\item We propose a self-supervised approach to segment drivable areas and road anomalies for robotic wheelchairs.
\item We develop a pipeline to automatically label drivable areas and road anomalies using RGB-D images.
\item We construct an RGB-D dataset\footnote{\url{https://github.com/hlwang1124/wheelchair-dataset.git}}, which covers 30 common scenes where robotic wheelchairs usually work.
\end{enumerate}

\section{Related Work}

\subsection{Drivable Area Segmentation}

Labayrade \textit{et al.} \cite{vdisparity} proposed an algorithm that converted the drivable area segmentation problem into a straight line detection problem. Following \cite{vdisparity}, some improvements\cite{gao2011uv, cong2010v, yiruo2013complex} were developed to enhance the robustness and accuracy. In recent years, some researchers have tried to solve this segmentation problem from other perspectives. For example,  Ozgunalp \textit{et al.} \cite{ozgunalp2016stereo} proposed to use the estimated planar patches and patch orientations to reduce the impact of outliers. Liu \textit{et al.} \cite{liu2018co} proposed an approach to fuse the information of light detection and ranging (LIDAR) and vision data, respectively.

\subsection{Road Anomaly Segmentation}

Early approaches of road anomaly segmentation mainly adopted traditional computer vision algorithms with handcrafted features. Cong \textit{et al.} \cite{dcmc} designed a feature based on the depth confidence analysis and multiple cues fusion. Lou \textit{et al.} \cite{smalltarget} proposed an approach combining both regional stability and saliency for small road anomaly segmentation.

With the development of deep learning, many work used deep neural networks to segment road anomalies. Peng \textit{et al.} \cite{nlpr} used a deep neural network to extract features from RGB-D images and fused these features in three levels to generate the predicted result. Chen \textit{et al.} \cite{chen2019multi} proposed a multi-scale multi-path fusion network with cross-modal interactions, which could explore deep connections between RGB images and depth images. There also exist some RGB-D data-based semantic segmentation neural networks that fuse RGB and depth data together such as FuseNet \cite{fusenet} and Depth-aware CNN \cite{depthaware}, which can achieve impressive performance. However, these methods rely on hand-labeled ground truth, which would become difficult to implement when there is no sufficient training dataset. Our proposed SSLG can automatically label drivable areas and road anomalies with RGB-D images, which greatly reduces the time and labor for manual labeling.

\begin{figure*}[t]
   \centering
   \includegraphics[width=\textwidth]{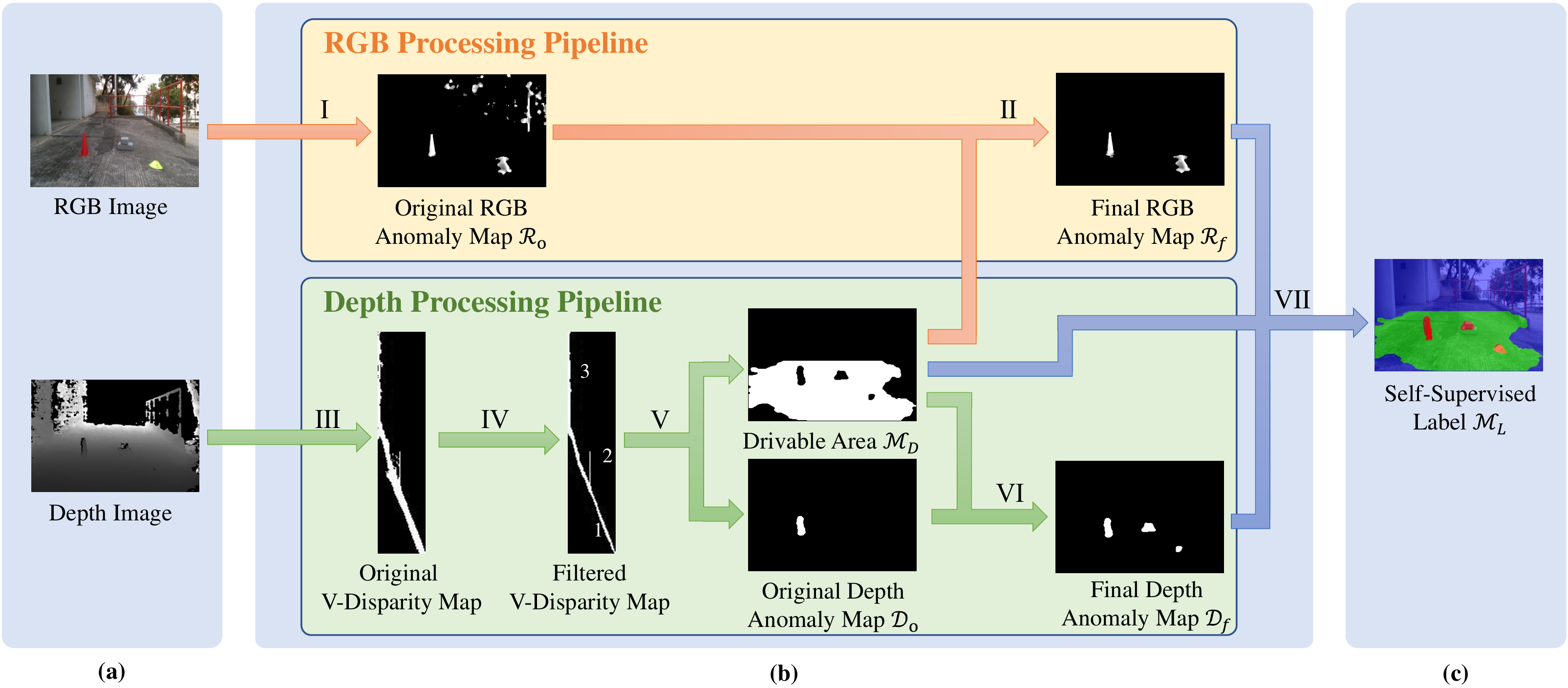}
   \caption{The overview of our proposed SSLG, which consists of (a) Input of RGB-D images, (b) Processing pipeline of RGB-D images and (c) Output of self-supervised labels. (b) is composed of RGB Processing Pipeline shown in the orange box, Depth Processing Pipeline shown in the green box and (\Rmnum{7}) Final Segmentation Label Generator shown in the blue lines. The RGB Processing Pipeline consists of (\Rmnum{1}) Original RGB Anomaly Map Generator and (\Rmnum{2}) Generation of final RGB anomaly maps. The Depth Processing Pipeline consists of (\Rmnum{3}) Computation of original v-disparity maps, (\Rmnum{4}) Filtering of original v-disparity maps, (\Rmnum{5}) Extraction of the drivable area and original depth anomaly maps as well as (\Rmnum{6}) Generation of final depth anomaly maps. The figure is best viewed in color.}
   \label{pipeline}
\end{figure*}

\subsection{Automatic Labeling}

Automatic labeling aims to generate labeling data automatically or in an unsupervised way without hand-labeled training data. Barnes \textit{et al.} \cite{barnes2017find} proposed an approach to generate training data for the segmentation problem of path proposals by exploiting the information of odometry and obstacle sensing. Mayr \textit{et al.} \cite{mayr2018self} designed a pipeline to automatically generate training data for the segmentation problem of drivable areas. However, neither of these two approaches can generate the label of drivable areas and road anomalies simultaneously.

With the development of graphic rendering engines, some researchers utilized the data from real-world environments to construct simulation environments, where it is effortless to acquire segmentation labels. Gaidon \textit{et al.} \cite{gaidon2016virtual} proposed a real-to-virtual world cloning approach to generate photo-realistic simulation environments. Xia \textit{et al.}  \cite{xia2018gibson} proposed Gibson Environment for developing real-world perception for active agents. However, there is a certain gap between the data collected in simulation environments and in real-world environments. For example, the depth images collected in simulation environments are very dense, but the actual depth images collected by RGB-D cameras have some invalid pixels. Besides, these simulation environments cannot cover the common indoor and outdoor scenes where robotic wheelchairs usually work and the road anomalies that robotic wheelchairs may encounter in real environments. Furthermore, in order to construct such a photo-realistic simulation environment for robotic wheelchairs, either large amounts of paired data for target-source domains or physical measurements of important objects in the scene are needed, which is time-consuming and labor-intensive.

\subsection{Self-Supervised Semantic Segmentation}
Zeng \textit{et al.} \cite{zeng2017multi} proposed a self-supervised approach to semantic segmentation by utilizing the known background information. However, their approach is only suitable for single confined scenes with objects that can be moved. In common scenes where robotic wheelchairs usually work, there are many anomalies (e.g., street lamps) that can not be moved. Besides, some researchers \cite{larsson2016learning, noroozi2016unsupervised, zhang2016colorful} designed proxy tasks (e.g., image colorization) to extract meaningful representations for self-supervised learning. Recently, Zhan \textit{et al.} \cite{zhan2018mix} presented a mix-and-match tuning approach for self-supervised semantic segmentation based on the existed proxy tasks.

\section{Proposed Method}

\subsection{Dataset Construction}

Note that we divide the common road anomalies for robotic wheelchairs into two categories: large road anomalies with a height larger than $15$cm from the surface of the drivable area; small road anomalies with a height between $5$cm to $15$cm from the surface of the drivable area. To the best of our knowledge, this is the first dataset that exhibit common road anomalies for robotic wheelchairs.

We use the Realsense camera to collect data involving both large and small road anomalies for the segmentation problem of robotic wheelchairs. Our dataset covers 30 common scenes where robotic wheelchairs usually work (e.g., sidewalks and squares) and 18 different kinds of road anomalies that robotic wheelchairs may encounter in real environments. Fig. 4 shows the number of finely annotated pixels for every kind of road anomaly in our dataset.

\begin{figure}[htbp]
   \centering
   \includegraphics[width=\linewidth]{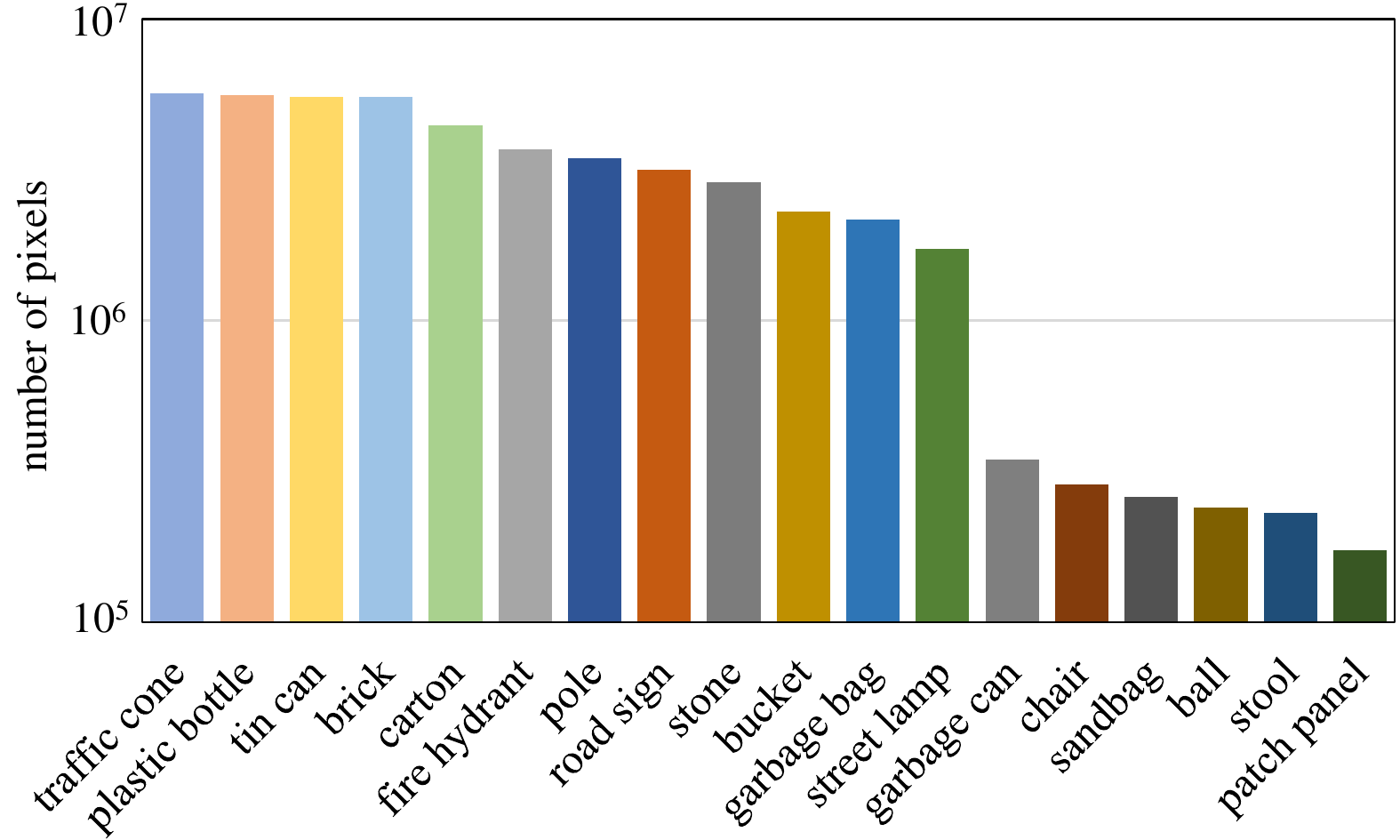}
   \caption{Number of finely annotated pixels (y-axis) and their associated categories (x-axis).}
   \label{anomalies}
   \vspace{-0.5cm}
\end{figure}

There are a total of 3896 RGB-D images with hand-labeled ground truth for segmentation in our dataset, which are with the image resolution of $720 \times 1280$ pixels. It should be noted that our proposed self-supervised approach does not require hand-labeled ground truth. The hand-labeled ground truth is only used for the evaluation of our proposed self-supervised approach. We provide two kinds of depth data in our dataset, the original depth data and the normalized depth data. The normalized depth data is normalized to the range of $0$ to $255$. Note that the distance measurement range for the Realsense RGB-D camera is up to $10$m. Therefore, we remove the pixels with the distance larger than $10$m and label them with the zero value. As for the hand-labeled ground truth, we label the unknown area with $0$, the drivable area with $1$ and road anomalies with $2$. The area except the drivable area and road anomalies is defined as the unknown area since it is not clear whether or not robotic wheelchairs can pass through it such as the area beyond the range of the Realsense camera.

\subsection{Self-Supervised Label Generator}
Our proposed Self-Supervised Label Generator (SSLG) is designed to generate self-supervised labels of drivable areas and road anomalies automatically. The overview of the SSLG is shown in Fig. 3.

We firstly elaborate the depth processing pipeline inspired by \cite{vdisparity}. As derived in \cite{vdisparity}, for an RGB-D camera consisting of two cameras, the projection of the real world point $P$ with coordinates of $(X, Y, Z)$ on the image coordinates $(U, V)$ can be computed by (1)--(3):

$$
U_{l}=u_{l}-u_{0}=f \frac{X + b / 2}{Y \sin \theta+Z \cos \theta}
\eqno{(1)}
$$

$$
U_{r}=u_{r}-u_{0}=f \frac{X - b / 2}{Y \sin \theta+Z \cos \theta}
\eqno{(2)}
$$

$$
V=v-v_{0}=f \frac{Y \cos \theta-Z \sin \theta}{Y \sin \theta+Z \cos \theta}
\eqno{(3)}
$$
where $b$ is the distance between the optical centers of two cameras; $f$ is the focal length; $(u_{0}, v_{0})$ is the center of the image plane; $u_{l}$, $u_{r}$ are the projection of the point $P$ on two cameras, respectively; $\theta$ is the pitch angle with respect to the ground plane. Then, the disparity $\Delta$ can be calculated by (4):
$$
\Delta=u_{l}-u_{r}=f \frac{b}{Y \sin \theta+Z \cos \theta}
\eqno{(4)}
$$

Horizontal planes in the real world coordinates can be represented by $Y = m$, which leads to:
$$
\Delta \frac{m}{b}=V \cos \theta+f \sin \theta
\eqno{(5)}
$$
Similarly, vertical planes in the real world coordinates can be represented by $Z = n$, which leads to:
$$
\Delta \frac{n}{b}=V \sin \theta+f \cos \theta
\eqno{(6)}
$$

Equation (5) and (6) show that horizontal planes and vertical planes in the real world coordinates can be projected as straight lines in the v-disparity map. Actually, \cite{vdisparity} proposed that this conclusion applies to all planes. The intuition behind the depth processing pipeline is that drivable areas can be regarded as planes in most cases and road anomalies can also be regarded as planes approximately. Then, the segmentation problem can be converted into a straight line extraction problem.

\begin{figure*}[t]
   \centering
   \includegraphics[width=\textwidth]{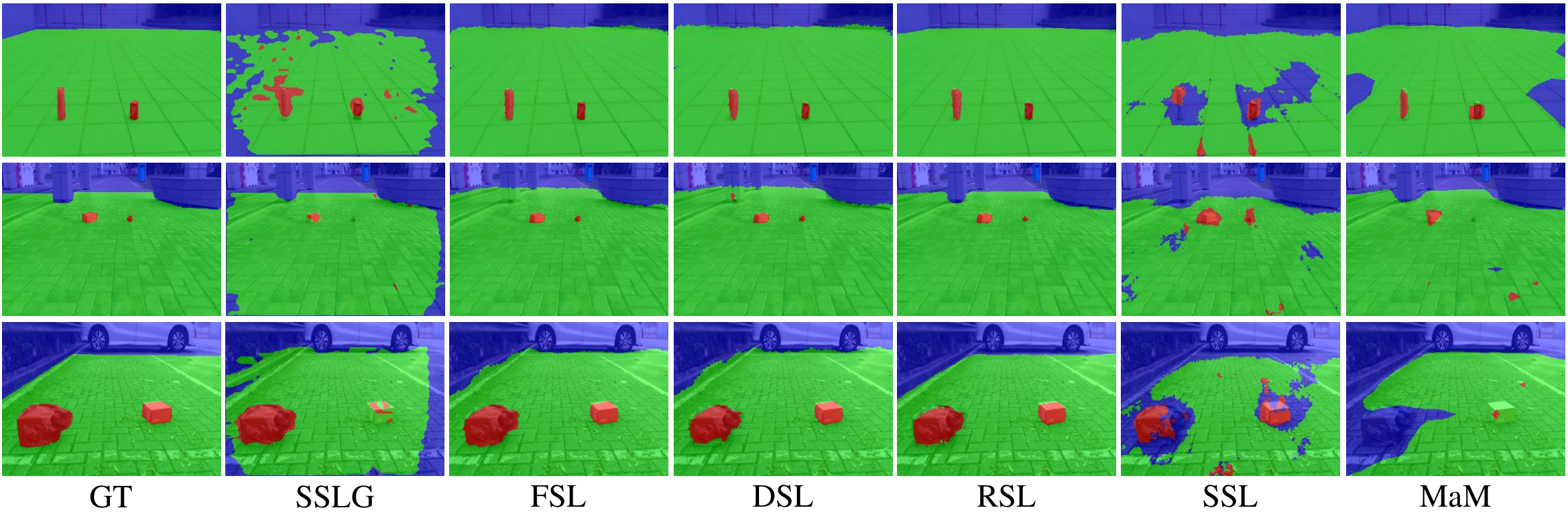}
   \caption{The comparison of the segmentation results between GT (manual labels), SSLG labels, FSL (FuseNet trained on the SSLG labels), DSL (Depth-aware CNN trained on the SSLG labels), RSL (RTFNet trained on the SSLG labels), SSL (SegNet trained on the SSLG labels with the input of only RGB images) and MaM (Mix-and-Match tuning for self-supervised semantic segmentation), where the blue area refers to the unknown area, the green area refers to the drivable area and the red area refers to road anomalies. The figure is best viewed in color.}
   \label{mainresult}
\end{figure*}

The original v-disparity map can be obtained by computing the depth histogram of each row in the depth image (Fig. 3 \Rmnum{3}). Since the computed v-disparity map often contains much noise, the steerable filter with the second derivatives of Gaussian as the basis function \cite{gao2011uv} is applied to filter the original v-disparity map (Fig. 3 \Rmnum{4}). Then, the Hough Transform algorithm \cite{Hough} is applied to extract straight lines in the filtered v-disparity map (Fig. 3 \Rmnum{4}). Gao \textit{et al.} \cite{gao2011uv} concluded that the drivable area is dominant in v-disparity maps; the straight line with the smallest disparity is the projection of the infinity plane; the remaining straight lines except the two straight lines mentioned above are marked as road anomalies. According to these conclusions, we firstly filter out the straight lines representing road anomalies with a height smaller than $5$cm from the surface of the drivable area according to their lengths. Then, it is easy to find that in the filtered v-disparity map, straight line $\text{No.}1$, $\text{No.}2$ and $\text{No.}3$ represent the drivable area, the road anomaly and the infinity plane, respectively. After that, we extract the drivable area $\mathcal{M}_{D}$ and the original depth anomaly map $\mathcal{D}_{o}$ according to the straight line detection results (Fig. 3 \Rmnum{5}).

However, the original depth anomaly map lacks robustness and accuracy because the straight lines representing small road anomalies are too short and easy to be filtered out together with the noise. For instance, there are three road anomalies in the example (Fig. 3), but there is only one straight line representing road anomalies in the filtered v-disparity map and thus one road anomaly detected in the original depth anomaly map. The other two small road anomalies (i.e., the brick and the road sign) are filtered out together with the noise. In order to solve this problem, we utilize the drivable area that we have already generated. We can find that there are some holes inside the drivable area, which contain the missing road anomalies in the original depth anomaly map. Therefore, we extract holes in the drivable area and then combine the hole detection results with the original depth anomaly map to generate the final depth anomaly map $\mathcal{D}_{f}$, which is further normalized to the range $[0, 1]$ (Fig. 3 \Rmnum{6}). Although this method will bring some noise to the depth anomaly map, it greatly increases the detection rate of road anomalies to ensure the safety of the riders and we will correct it again with the information of RGB images.

\begin{algorithm}[htbp]
   \providecommand{\DontPrintSemicolon}{\dontprintsemicolon}
   \KwIn{$\mathcal{L}$, $\mathbf{h}$, $\mathbf{w}$, $\sigma_{s}$.}
   \KwOut{$\mathcal{R}_{o}$.}
   $\sigma=\min(\mathbf{h}, \mathbf{w})/\sigma_{s}$ \\
   initialize $\mathcal{L}_{\omega}$ with three channels $(l_{\omega}, a_{\omega}, b_{\omega})$\\
   construct a Gaussian kernel $\mathcal{G}$ with the size $3\sigma\times3\sigma$ and the standard deviation $\sigma$ \\
   $\mathcal{L}_{\omega} = \mathcal{G}(\mathcal{L})$ \\
   $\mathcal{R}_{o} = \left\|\mathcal{L}-\mathcal{L}_{\omega}\right\|^{2}$
   \caption{Original RGB Anomaly Map Generator}
   \label{algorithm1}
\end{algorithm}

Now we elaborate the RGB image processing pipeline inspired by \cite{smalltarget}. The intuition behind the RGB processing pipeline is that the areas with different colors from surrounding areas are often marked as road anomalies. Based on this principle, we design an original RGB anomaly map generator (Fig. 3 \Rmnum{1}), which is described in Algorithm $1$. Let $\mathcal{L}$ denote the image in the $Lab$ color space transformed from the RGB image, and $\mathbf{h}$ and $\mathbf{w}$ denote the width and the height of the RGB image, respectively. We generate the original RGB anomaly map $\mathcal{R}_{o}$ by computing the difference between the $Lab$ color vector of each pixel and its Gaussian blurred result. In order to suppress the pattern of the drivable area, we choose a large filter scale for the Gaussian kernel to blur each channel of the $Lab$ color space. The size of the kernel is $3\sigma\times3\sigma$ with the standard deviation $\sigma$ and $\sigma_{s}$ is chosen to be $12$ to control the strength of weighting.

However, the original RGB anomaly map lacks robustness and accuracy because of the interference outside the drivable area. In order to solve this problem, we utilize the drivable area that we have already generated to filter out the noise outside the drivable area (Fig. 3 \Rmnum{2}). After normalized to the range $[0, 1]$, the final RGB anomaly map $\mathcal{R}_{f}$ is generated.

The last step of our proposed SSLG is to combine two anomaly maps and the drivable area to generate the self-supervised label. We design a final segmentation label generator (Fig. 3 \Rmnum{7}), which is described in Algorithm $2$. As for road anomalies, we firstly generate the final anomaly map $\mathcal{M}_{A}$ according to (7):
$$
\mathcal{M}_{A} = \alpha \mathcal{R}_{f} + (1-\alpha) \mathcal{D}_{f}
\eqno{(7)}
$$
Then, we set a threshold $\kappa$ and the area in the final anomaly map where the value is greater than $\kappa$ is marked as road anomalies in the self-supervised label. In our case, we set $\alpha$ to $0.5$ and $\kappa$ to $0.3$. The drivable area in the self-supervised label is the same as the drivable area $\mathcal{M}_{D}$, and the rest area except drivable areas and road anomalies marked above is labeled as the unknown area. Finally the self-supervised label $\mathcal{M}_{L}$ is generated, which is used for training RGB-D data-based semantic segmentation neural networks as described in the following sections.

\subsection{RGB-D Data-based Semantic Segmentation Neural Networks}

We use our proposed SSLG to generate self-supervised labels for a total of 3896 images in our constructed dataset, which can then be utilized to train RGB-D data-based semantic segmentation neural networks. Here, we use three off-the-shelf RGB-D data-based semantic segmentation neural networks, FuseNet \cite{fusenet}, Depth-aware CNN \cite{depthaware} and RTFNet \cite{rtfnet}. Note that RTFNet is initially designed for RGB and thermal data fusion, but we find that it generalizes well to RGB-D data. We also use one off-the-shelf RGB data-based semantic segmentation neural network SegNet \cite{segnet} for ablation study. The total 3896 images are split into the training set, the validation set and the test set that contains 2726, 585 and 585 images, respectively. Each set contains different scenes and anomalies from the other two sets. Each network is trained on the training set with the SSLG labels. Note that we also train each network on the training set with manual labels for the evaluation of our proposed self-supervised approach.

\begin{algorithm}[t]
   \providecommand{\DontPrintSemicolon}{\dontprintsemicolon}
   \KwIn{$\mathcal{R}_{f}$, $\mathcal{D}_{f}$, $\mathcal{M}_{D}$, $\mathbf{h}$, $\mathbf{w}$, $\kappa$.}
   \KwOut{$\mathcal{M}_{L}$.}
   compute $\mathcal{M}_{A}$ using (7) \\
   \For{$i \gets 1$ \textbf{t o} $\mathbf{h}$}
   {
      \For{$j \gets 1$ \textbf{to} $\mathbf{w}$}
      {
         \uIf{$\mathcal{M}_{A}(i,j) > \kappa$}
         {
            label $\mathcal{M}_{L}(i,j)$ as road anomalies
         }
         \uElseIf{$\mathcal{M}_{D}(i, j)$ \rm{is labeled positive}}
         {
            label $\mathcal{M}_{L}(i,j)$ as the drivable area
         }
         \Else{  label $\mathcal{M}_{L}(i,j)$ as the unknown area}
      }
   }
   \caption{Final Segmentation Label Generator}
   \label{algorithm2}
\end{algorithm}

\begin{table*}[h]
	\caption{The comparison of the segmentation results ($\%$) between SSLG labels, FSL (FuseNet trained on the SSLG labels), FML (FuseNet trained on the manual labels), DSL (Depth-aware CNN trained on the SSLG labels), DML (Depth-aware CNN trained on the manual labels), RSL (RTFNet trained on the SSLG labels), DML (RTFNet trained on the manual labels), SSL (SegNet trained on the SSLG labels with the input of only RGB images), SML (SegNet trained on the manual labels with the input of only RGB images) and MaM (Mix-and-Match tuning for self-supervised semantic segmentation). Best results without using manual labels are highlighted in bold font.}
	\centering
	\begin{tabular}{C{1.5cm}C{0.9cm}C{0.9cm}C{0.9cm}C{0.9cm}C{0.9cm}C{0.9cm}C{0.9cm}C{0.9cm}C{0.9cm}C{0.9cm}C{0.9cm}C{0.9cm}}
	\toprule
	\multicolumn{1}{c}{\multirow{2}{*}{Approach}} & \multicolumn{3}{c}{Unknown Area} & \multicolumn{3}{c}{Drivable Area} & \multicolumn{3}{c}{Road Anomalies} & \multicolumn{3}{c}{All} \\ \cmidrule(l){2-4} \cmidrule(l){5-7} \cmidrule(l){8-10} \cmidrule(l){11-13}
	\multicolumn{1}{c}{} & Pre & Rec & IoU & Pre & Rec & IoU & Pre & Rec & IoU & Pre & Rec & IoU \\ \midrule
	SSLG & 89.62 & 80.36 & 75.09 & 75.70 & 86.92 & 65.87 & 33.15 & 22.92 & 16.03 & 66.16 & 63.40 & 52.33 \\ \midrule
	FSL & 82.76 & \textbf{91.52} & 78.26 & \textbf{88.19} & 77.72 & 75.22 & \textbf{75.57} & 64.94 & \textbf{54.36} & \textbf{82.17} & 78.06 & 69.28 \\
	FML & 82.80 & 99.86 & 82.70 & 99.60 & 82.15 & 81.82 & 90.14 & 77.22 & 71.20 & 90.85 & 86.41 & 78.57 \\
	DSL & 78.25 & 88.63 & 79.12 & 84.57 & 78.97 & 74.96 & 65.11 & \textbf{72.57} & 48.93 & 75.98 & 80.06 & 67.67 \\
	DML & 79.58 & 99.91 & 79.52 & 99.65 & 79.48 & 79.25 & 86.76 & 73.95 & 66.45 & 88.66 & 84.45 & 75.07 \\
	RSL & \textbf{90.98} & 86.81 & \textbf{83.62} & 83.86 & \textbf{90.12} & \textbf{82.70} & 60.66 & 71.89 & 50.46 & 78.50 & \textbf{82.94} & \textbf{72.26} \\
	RML & 95.03 & 94.71 & 92.77 & 95.84 & 95.26 & 92.14 & 72.35 & 85.49 & 66.09 & 87.74 & 91.82 & 83.67 \\ \midrule
	SSL & 83.92 & 81.31 & 69.20 & 70.37 & 75.97 & 70.47 & 61.92 & 52.28 & 42.91 & 72.07 & 69.85 & 60.86 \\
	SML & 96.97 & 90.32 & 80.30 & 84.49 & 99.37 & 77.61 & 80.91 & 55.79 & 50.42 & 87.46 & 81.83 & 69.44 \\
	MaM & 78.77 & 87.91 & 73.85 & 87.40 & 75.92 & 71.62 & 62.25 & 68.84 & 41.12 & 76.14 & 77.56 & 62.20 \\ \bottomrule
	\end{tabular}
\end{table*}

\section{Experimental Results and Discussions}

In this section, we evaluate the performance of the segmentation results on our constructed dataset and compare our proposed self-supervised approach with other state-of-the-art methods.

\subsection{Segmentation Results}

As aforementioned, FuseNet, Depth-aware CNN, RTFNet and SegNet are all trained on the training set in our constructed dataset with the SSLG labels and manual labels, respectively. For all networks, the stochastic gradient descent (SGD) and a base learning rate of $0.001$ is used. Note that the resolution of input images is downsampled to $480 \times 640$. Each network is trained for 400 epochs and then selected the best model according to the performance of the validation set. After obtaining the best models, we utilize the manual labels of 585 images in our constructed test set to evaluate the performance of these models. Fig. 5 shows the comparison of the segmentation results for three images. As aforementioned, we do not take the size of the robotic wheelchair into consideration, because planning algorithms would use this information to determine whether the vehicle can pass through. From the figure, we can see that the SSLG labels miss some road anomalies, while three RGB-D networks trained on the SSLG labels can detect all road anomalies, which demonstrates that our proposed self-supervised approach can learn the features of road anomalies and drivable areas from the noisy SSLG labels. These results are also backed up by the quantitative evaluation. Tab. \Rmnum{1} presents the evaluation results for three segmentation classes, the unknown area, the drivable area and road anomalies, where Pre, Rec and IoU are short for precision, recall and intersection-over-union, respectively. The "All" column shows the mean of precision, recall and IoU of all three segmentation classes, respectively. It is obvious that three RGB-D networks trained on the SSLG labels present certain improvements compared to the SSLG labels trained on the SSLG labels in the unknown area and the drivable area. As for road anomalies, three RGB-D networks trained on the SSLG labels present significant improvements compared to the SSLG labels, which leads to the obvious mean IoU improvement of $16.95\%$, $15.34\%$ and $19.93\%$ in the "All" column for FuseNet, Depth-aware CNN and RTFNet respectively. The reason why our proposed self-supervised approach exhibits more robust and accurate results than the SSLG labels is that our proposed SSLG is based on the intuition that drivable areas and road anomalies can be regarded as planes approximately, which is deviated slightly from the real environments. However, our proposed self-supervised approach is capable to implicitly learn useful and effective features of drivable areas and road anomalies from huge amounts of contradicting labels.

\begin{figure}[t]
   \centering
   \includegraphics[width=\linewidth]{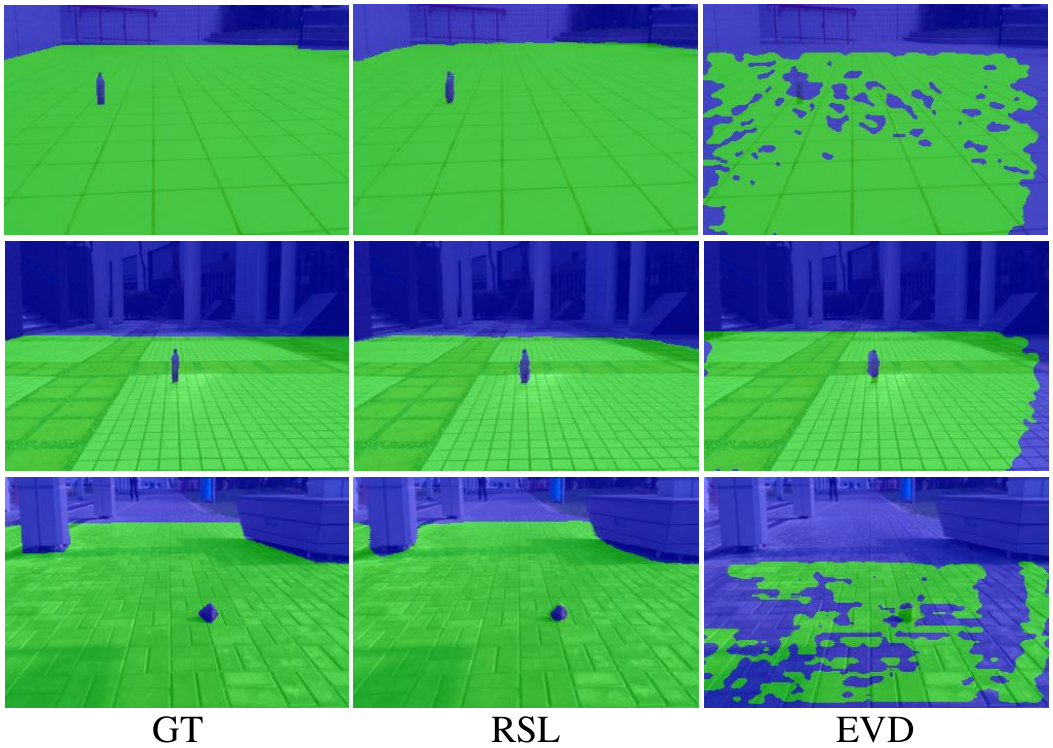}
   \caption{The comparison of drivable area segmentation results between GT (manual labels), RSL (RTFNet trained on the SSLG labels) and EVD\cite{yiruo2013complex}, where the green area refers to the drivable area and the blue area refers to other areas. The figure is best viewed in color.}
   \label{roadresult}
\end{figure}

\begin{figure*}[htbp]
   \centering
   \includegraphics[width=\textwidth]{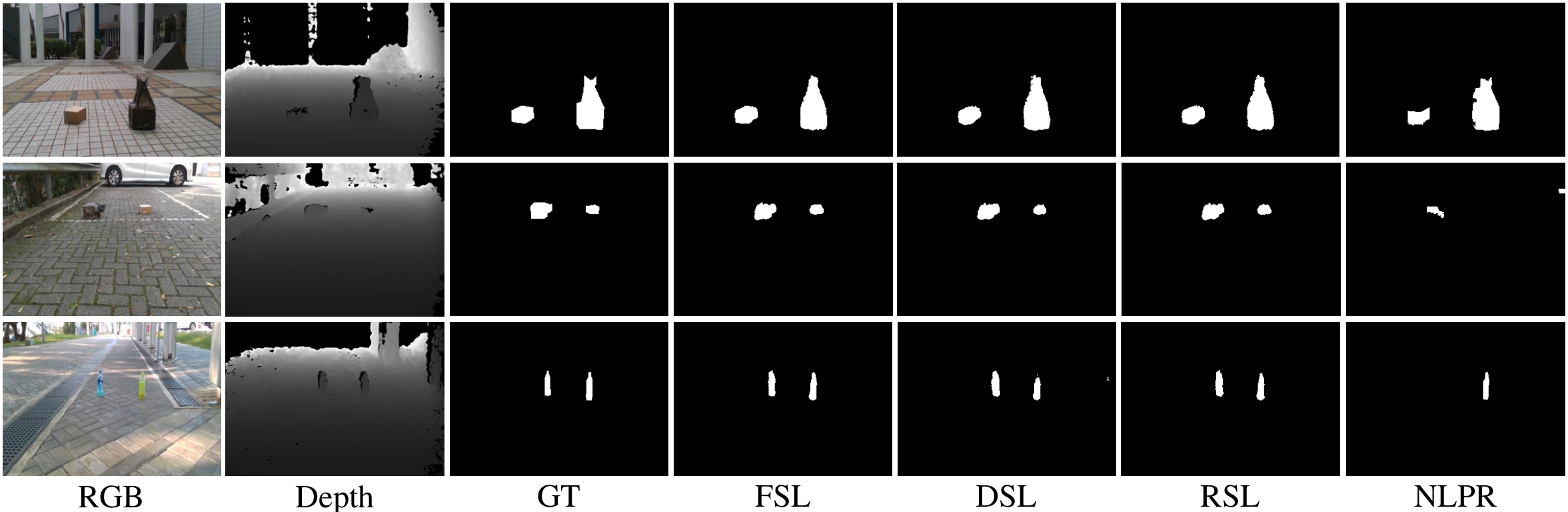}
   \caption{The comparison of road anomaly segmentation results between NLPR\cite{nlpr}, FSL (FuseNet trained on the SSLG labels), DSL (Depth-aware CNN trained on the SSLG labels) and RSL (RTFNet trained on the SSLG labels).}
   \label{anomalyresult}
\end{figure*}

As expected, three RGB-D networks present better performance when trained on the manual labels. However, it requires much time and intensive labor to create a dataset with manually-labeled ground truth. According to the experiments, our proposed SSLG only takes 2 seconds to label one image. Although our proposed segmentation method introduces more errors, it is still remarkable considering the impressive speed-up compared to manual labeling with the achievable results.

\subsection{Ablation Study}

As aforementioned, we use one off-the-shelf RGB data-based semantic segmentation neural network SegNet \cite{segnet} trained on the SSLG labels (SSL) for ablation study. Fig. 5 shows the comparison of the segmentation results for three images. From the figure, we can see that three RGB-D networks trained on the SSLG labels present a better performance than SegNet trained on the SSLG labels. The quantitative results shown in Tab. \Rmnum{1} also confirms our conclusion that compared to SSL, our proposed FSL has a mean precision improvement of $10.10\%$ and our proposed RSL has a recall improvement of $13.09\%$ as well as an IoU improvement of $11.40\%$. The reason is that the depth information of road anomalies and drivable ares are really useful to distinguish them.

\subsection{Comparison with Other State-of-the-art Methods}

In order to demonstrate the robustness and accuracy of our proposed self-supervised approach, we compare its performance with other state-of-the-art methods.

\textit{1) Self-Supervised Semantic Segmentation:}
Zhan \textit{et al.} \cite{zhan2018mix} presented a mix-and-match tuning approach (MaM), a state-of-the-art approach for self-supervised semantic segmentation. We test this approach on the 585 images in our constructed test set and compare its performance with FuseNet trained on the SSLG labels (FSL), Depth-aware CNN trained on the SSLG labels (DSL) and RTFNet trained on the SSLG labels (RSL). Fig. 5 presents the segmentation results of three images, from which we can see that our proposed self-supervised approach presents a great advantage over MaM. The quantitative results shown in Tab. \Rmnum{1} also confirms our conclusion that compared to MaM, our proposed FSL has a mean precision improvement of $6.03\%$ and our proposed RSL has a recall improvement of $5.38\%$ as well as an IoU improvement of $10.06\%$.

\textit{2) Drivable Area Segmentation:}
Yiruo \textit{et al.} \cite{yiruo2013complex} proposed EVD, a state-of-the-art method of the drivable area segmentation based on the enhanced v-disparity map. We use this method on the 585 images in our constructed test set and compare its performance with FuseNet trained on the SSLG labels (FSL) and Depth-aware CNN trained on the SSLG labels (DSL). Fig. 6 presents the segmentation results of three images, from which we can see that our proposed self-supervised approach presents a great advantage over EVD. The quantitative results shown in Tab. \Rmnum{2} also confirms our conclusion that compared to EVD, our proposed FSL has a precision improvement of $2.83\%$ and our proposed RSL has a recall improvement of $33.30\%$ as well as an IoU improvement of $30.50\%$.

\begin{table}[t]
\centering
\caption{The comparison of drivable area segmentation results ($\%$) between EVD\cite{yiruo2013complex}, FSL (FuseNet trained on the SSLG labels), DSL (Depth-aware CNN trained on the SSLG labels) and RSL (RTFNet trained on the SSLG labels). Best results are highlighted in bold font.}
\begin{tabular}{@{}C{2cm}C{1.8cm}C{1.8cm}C{1.8cm}@{}}
\toprule
Approach & Precision & Recall & IoU \\ \midrule
EVD & 85.36 & 56.82 & 52.20 \\
FSL & \textbf{88.19} & 77.72 & 75.22 \\
DSL & 84.57 & 78.97 & 74.96 \\
RSL & 83.86 & \textbf{90.12} & \textbf{82.70} \\ \bottomrule
\end{tabular}
\end{table}

\begin{table}[t]
   \centering
   \caption{The comparison of road anomaly segmentation results ($\%$) between NLPR\cite{nlpr}, FSL (FuseNet trained on the SSLG labels), DSL (Depth-aware CNN trained on the SSLG labels) and RSL (RTFNet trained on the SSLG labels). Best results are highlighted in bold font.}
   \begin{tabular}{@{}C{2cm}C{1.8cm}C{1.8cm}C{1.8cm}@{}}
   \toprule
   Approach & Precision & Recall & IoU \\ \midrule
   NLPR & 43.62 & 49.27 & 30.36 \\
   FSL & \textbf{75.57} & 64.94 & \textbf{54.36} \\
   DSL & 65.11 & \textbf{72.57} & 48.93 \\
   RSL & 60.66 & 71.89 & 50.46 \\ \bottomrule
   \end{tabular}
\end{table}

\textit{3) Road Anomaly Segmentation:}
As for the problem of road anomaly segmentation, we applied NLPR\cite{nlpr} on the 585 images in our constructed test set. Then, we filter out the anomalies segmentation results outside the drivable area by utilizing the manual labels to ensure the fairness and rationality of the comparison results. Fig. 7 presents the segmentation results of three images, from which we can see that our proposed self-supervised approach presents a great advantage over NLPR. The quantitative results shown in Tab. \Rmnum{3} also confirms our conclusion that compared to NLPR, our proposed DSL has a recall improvement of $23.30\%$ and our proposed FSL has a precision improvement of $31.95\%$ as well as an IoU improvement of $24.00\%$.

\begin{figure}[t]
  \begin{subfigure}[t]{0.49\linewidth}
    \centering
    \includegraphics[width=\linewidth]{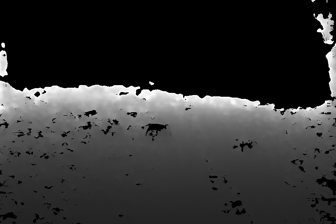}
    \caption{Depth image}
  \end{subfigure}
  \hfill
  \begin{subfigure}[t]{0.49\linewidth}
    \centering
    \includegraphics[width=\linewidth]{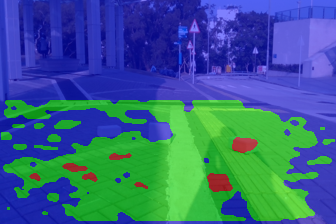}
    \caption{Segmentation of SSLG labels}
  \end{subfigure}

  \begin{subfigure}[t]{0.49\linewidth}
    \centering
    \includegraphics[width=\linewidth]{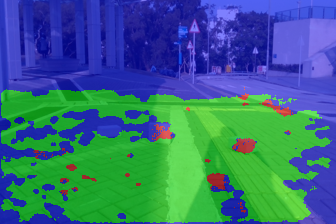}
    \caption{Segmentation of FSL}
  \end{subfigure}
  \hfill
  \begin{subfigure}[t]{0.49\linewidth}
    \centering
    \includegraphics[width=\linewidth]{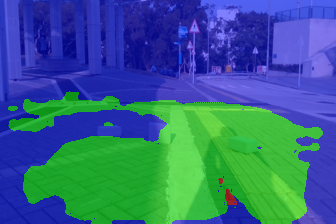}
    \caption{Segmentation of DSL}
  \end{subfigure}
  \caption{Typical failure cases due to the noise in the depth image. The figure is best viewed in color.}
  \label{limitations}
\end{figure}

\subsection{Limitations}
In some cases, the RGB-D data-based semantic segmentation neural networks fail to output stable and accurate segmentation results, as illustrated in Fig. 8. This is because some depth images have much noise, which leads to the inaccurate SSLG labels and thus the inaccurate segmentation results. This problem can be solved by using high-quality RGB-D cameras.

\section{Conclusions}

In this paper, we presented a comprehensive study on the drivable area and road anomaly segmentation problem for robotic wheelchairs. A self-supervised approach was proposed, which contains an automatic labeling pipeline for drivable area and road anomaly segmentation. Experimental results showed that our proposed automatic labeling pipeline achieved an impressive speed-up compared to manual labeling. In addition, our proposed self-supervised approach exhibited more robust and accurate results than the state-of-the-art traditional algorithms as well as the state-of-the-art self-supervised algorithms. In our future work, we plan to investigate our work with planning algorithms for robotic wheelchairs to achieve autonomous navigation.


\end{document}